\documentclass[conference,a4paper]{APSIPA2025}
\usepackage[T1]{fontenc}
\usepackage{amsmath}
\usepackage{graphicx}
\usepackage{multirow}
\usepackage{threeparttable}
\usepackage[backend=biber,style=ieee,]{biblatex}
\usepackage{booktabs}
\usepackage{pifont}
\newcommand{\cmark}{\ding{51}}  % checkmark
\newcommand{\xmark}{\ding{55}}  % cross
\usepackage{geometry}
\usepackage{fancyhdr}

\fancypagestyle{firststyle}{
  \fancyhf{}
  \fancyhead[C]{2025 Asia Pacific Signal and Information Processing Association Annual Summit and Conference (APSIPA ASC)}
}

\begin{document}

\title{KH-FUNSD: A Hierarchical and Fine-Grained Layout Analysis Dataset for Low-Resource Khmer Business Document}

\author{
\authorblockN{
Nimol Thuon\authorrefmark{1} and
Jun Du\authorrefmark{1}\textsuperscript{\dag}
}

\authorblockA{
\authorrefmark{1}
National Engineering Research Center of Speech and Language Information Processing (NERC-SLIP)\\ University of Science and Technology of China, Hefei, Anhui, 230027, China \\
E-mail: tnimol@mail.ustc.edu.cn, jundu@ustc.edu.cn}
}

\maketitle
\renewcommand{\thefootnote}{\fnsymbol{footnote}}
\footnotetext{\textsuperscript{$\dag$}Corresponding author}
\renewcommand{\thefootnote}{\arabic{footnote}}
\thispagestyle{firststyle}
\pagestyle{fancy}

\begin{abstract}
Automated document layout analysis remains a major challenge for low-resource, non-Latin scripts. Khmer is a language spoken daily by over 17 million people in Cambodia, receiving little attention in the development of document AI tools. The lack of dedicated resources is particularly acute for business documents, which are critical for both public administration and private enterprise. To address this gap, we present \textbf{KH-FUNSD}, the first publicly available, hierarchically annotated dataset for Khmer form document understanding, including receipts, invoices, and quotations. Our annotation framework features a three-level design: (1) region detection that divides each document into core zones such as header, form field, and footer; (2) FUNSD-style annotation that distinguishes questions, answers, headers, and other key entities, together with their relationships; and (3) fine-grained classification that assigns specific semantic roles, such as field labels, values, headers, footers, and symbols. This multi-level approach supports both comprehensive layout analysis and precise information extraction. We benchmark several leading models, providing the first set of baseline results for Khmer business documents, and discuss the distinct challenges posed by non-Latin, low-resource scripts. The KH-FUNSD dataset and documentation will be available at \url{https://github.com/back-kh/KH-FUNSD}.

\end{abstract}

\section{Introduction}

The extraction of structured information from form-like documents is fundamental to digital transformation across sectors \cite{funsd,xfund,vu2020revising}. Automated document layout analysis, understanding both the spatial and semantic structure of documents such as receipts and invoices, enables efficient data entry, large-scale digitization, business analytics, and regulatory compliance. Recent advances in layout-aware deep learning models, such as LayoutLM and its successors, have significantly improved performance in this field \cite{layoutlm1,layoutlmv2, LayoutLMv3}. However, these models and their associated benchmarks have been developed primarily for high-resource, Latin-script languages, leaving non-Latin and low-resource scripts underrepresented.

\begin{figure*}[h!]
\begin{center}     
\includegraphics[width=0.7\textwidth]{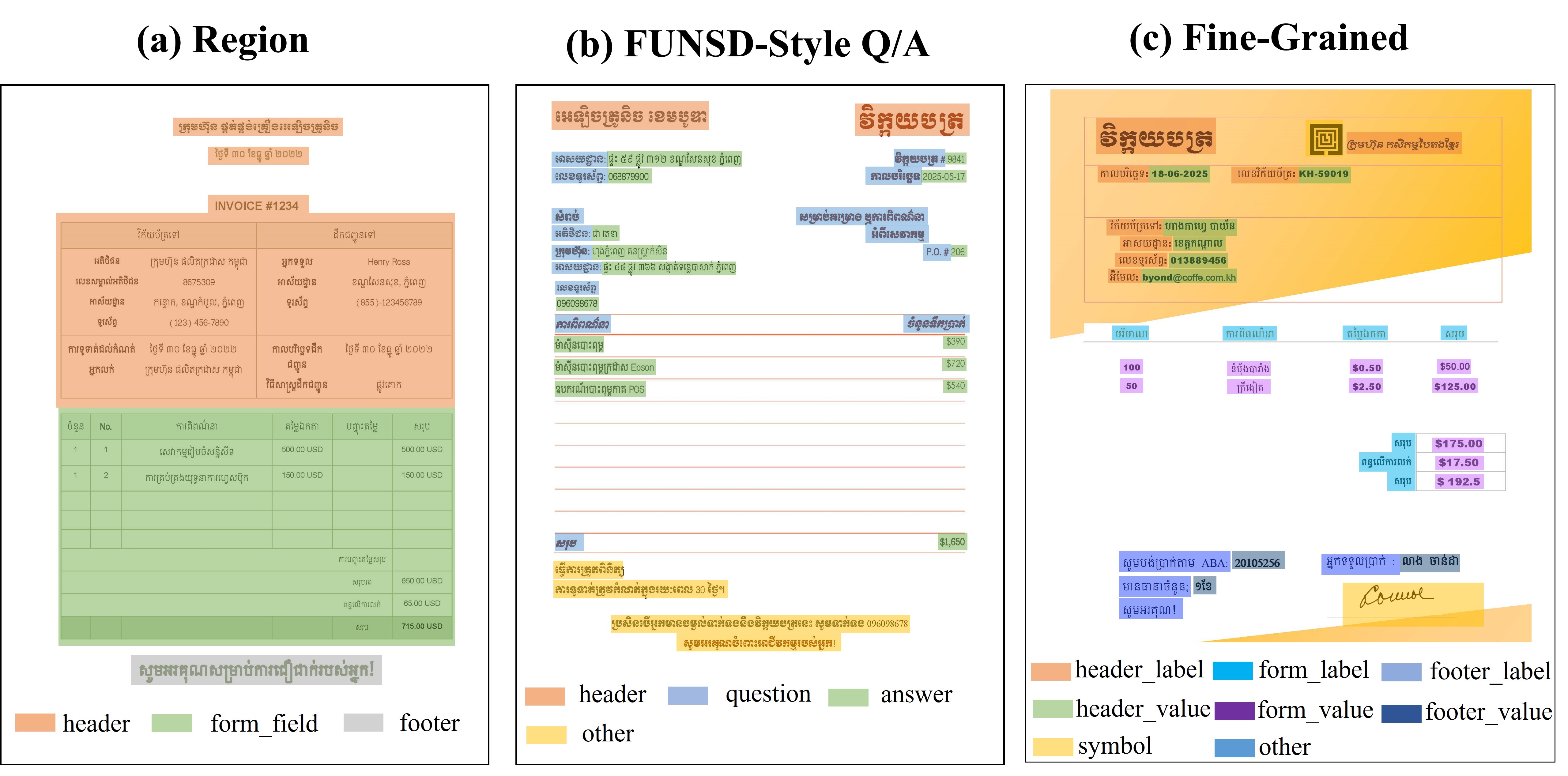}
\caption{Example annotated Khmer business documents: (a) Region-level detection identifies zones such as header, form field, and footer; (b) FUNSD-style Q/A annotation assigns roles including header, question, answer, and other; (c) Fine-grained classification provides explicit label–value pairs across all regions, including headers, form fields, footers, symbols and other.}
\label{fig:1}
\end{center}
\end{figure*}

Khmer, the official and most widely spoken language of Cambodia, is used daily by over 17 million people \cite{7ModernKhmer,Khmer, thuon2024khmerformer}. As Cambodia undergoes rapid digitalization, the demand for Khmer-language AI tools and resources has become increasingly urgent. Business documents such as receipts, invoices, and quotations play a critical role in commerce, public administration, taxation, and archival systems, yet remain largely underserved by existing document analysis technologies \cite{kabak2010survey}. Most current tools and datasets are tailored to English or other high-resource languages, making it challenging to directly apply state-of-the-art models to Khmer. The script itself presents unique challenges, including complex ligatures, stacked characters, and the lack of whitespace between words, all of which significantly hinder OCR and layout analysis \cite{thuon2024generate,THUON20258}. Furthermore, the absence of publicly available annotated datasets for Khmer business documents further limits both research progress and practical deployment \cite{thuon2022syllablekhmer,thuon2022improving}.

To address these challenges, we introduce \textbf{KH-FUNSD}, the first publicly available, hierarchically annotated dataset and evaluation benchmark for Khmer business documents. Our annotation scheme adopts a three-level framework, consisting of region-level detection, FUNSD-style entity linking, and fine-grained semantic labeling, which together capture both structural and semantic aspects of business documents.

The main contributions of this work are as follows:
\begin{itemize}
    \item We introduce \textbf{KH-FUNSD}, the first dataset and benchmark for Khmer Form Document Understanding, with comprehensive region annotations, FUNSD-style Q\&A, and hierarchical annotations for layout analysis.
    \item We provide baseline evaluations using state-of-the-art models (YOLO, DETR, LayoutLM) for region detection and semantic role prediction.
    \item We make all dataset, guidelines publicly available to support further research and development in this field.
\end{itemize}

This multi-level annotation approach enables robust structural layout modeling and precise semantic information extraction. By providing the first standardized evaluation for Khmer receipts and invoices, our work aims to bridge the gap in low-resource document AI and promote inclusive research for underrepresented languages across Southeast Asia.

\begin{figure*}[h!]
\begin{center}     
\includegraphics[width=0.7\textwidth]{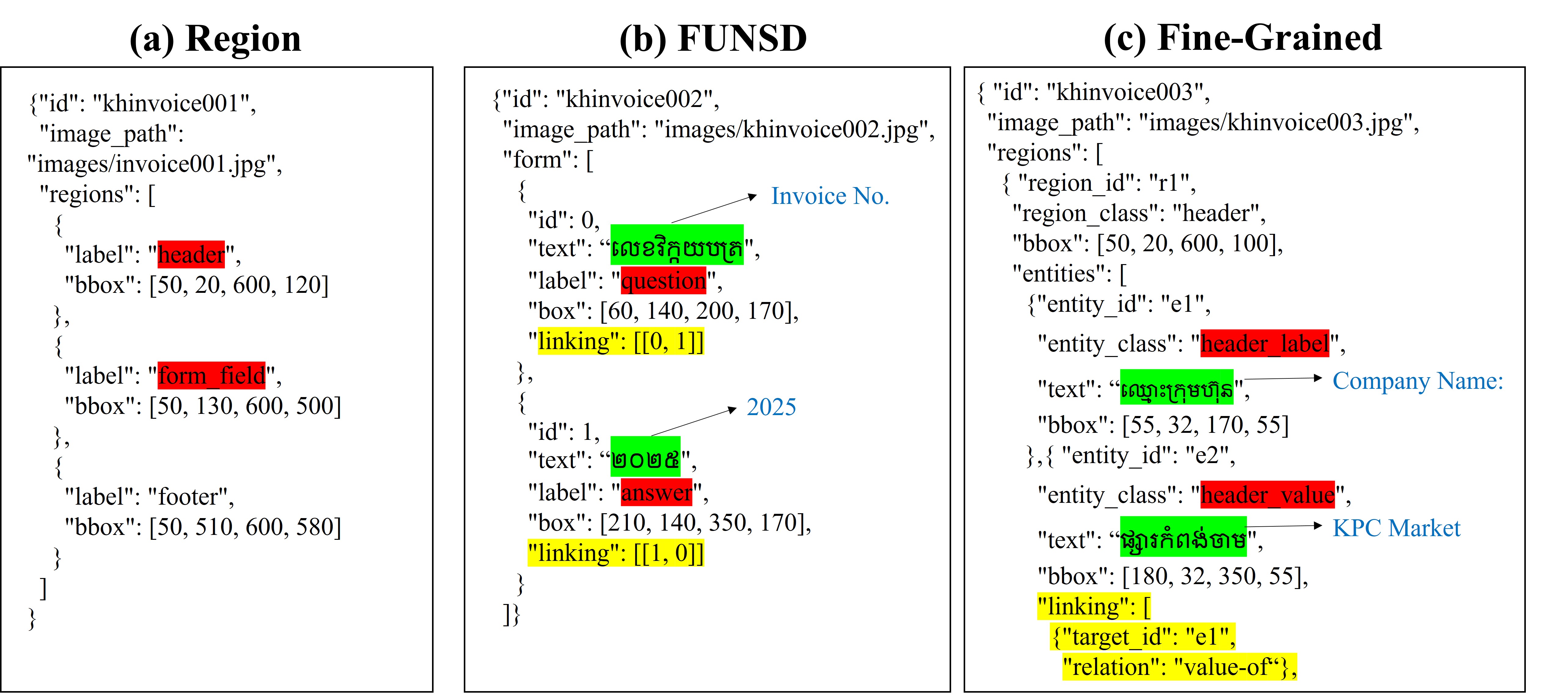}
\caption{Example JSON annotation formats: (a) Region-level detection, (b) FUNSD-style Q/A annotation, and (c) fine-grained label–value annotation. In the illustration, label classes are highlighted in red, label values in green, and links between components are shown in yellow.}
\label{fig:2}
\end{center}
\end{figure*}

\section{Related Work}

\subsection{Layout Analysis and Document Understanding}

Document layout analysis has evolved from early rule-based systems to modern deep learning models that integrate both textual and visual features. In high-resource languages, particularly English-models such as LayoutLM \cite{layoutlm1}, LayoutLMv2 \cite{layoutlmv2}, and LayoutLMv3 \cite{LayoutLMv3} have achieved state-of-the-art performance in tasks like region detection, semantic labeling, and key-value information extraction. These advances have been driven by the availability of large annotated datasets, high-quality OCR engines, and continued research focus.

However, such progress has not extended equally to low-resource and non-Latin scripts, which face challenges including complex writing systems, variable document layouts, and a lack of labeled training data. While recent efforts have begun to address Chinese \cite{chinesedataset}, Arabic \cite{arabicdataset}, and Japanese \cite{Japanesedataset} through specialized datasets and model adaptations, Southeast Asian scripts such as Khmer remain largely overlooked. The unique visual and linguistic characteristics of these scripts, combined with the absence of annotated resources, limit the development and evaluation of layout-aware models. Cross-lingual transfer and script adaptation techniques offer some promise, but their effectiveness is constrained by the lack of domain-specific training data. In parallel, newer models like DETR \cite{DETR} and Donut \cite{kim2022ocr} have explored OCR-free and multi-modal approaches for end-to-end document understanding, though these models have been primarily developed for high-resource, Latin-based scripts and are yet to be validated in low-resource, non-Latin contexts.

\subsection{Existing Datasets and Benchmarks}

Several benchmark datasets have significantly advanced the field of document understanding, particularly for Latin-script documents. The FUNSD \cite{funsd} and XFUND \cite{xfund} datasets are widely used for form understanding and key-value extraction, providing annotations for semantic entity types and inter-entity relationships. RVL-CDIP \cite{van2024beyond} offers a large-scale corpus of scanned document images across sixteen categories and is frequently employed for document classification and high-level layout analysis. The SROIE \cite{huang2019icdar2019} dataset, introduced for the ICDAR 2019 competition, focuses on receipt understanding, including named entity labeling and key-value extraction. CORD \cite{park2019cord} further extends this task to multilingual settings, particularly Korean and English, with rich layout and OCR annotations. Recently, datasets targeting Khmer scene text detection and recognition have also been introduced, which support script-specific Khmer detection and recognition tasks \cite{nom2024khmerst}. Despite their contributions, these datasets primarily address high-resource, Latin-based scripts and do not capture the diversity or structural complexity of low-resource scripts or Southeast Asian documents. Few benchmarks exist for comprehensive document understanding in non-Latin scripts, and even fewer provide hierarchical or fine-grained region-level annotations necessary for structured information extraction. This gap remains a significant barrier to developing inclusive and scalable document AI systems for underrepresented languages such as Khmer.

\section{Dataset Construction}

To advance research in document layout analysis and information extraction for Khmer business documents, we introduce a new annotated dataset comprising printed and scanned receipts, invoices, and quotations. The construction process was designed to ensure representative coverage of real-world Khmer layouts while fully preserving data privacy. The dataset integrates documents from open-source sources as well as manually synthesized samples inspired by authentic business templates, with all sensitive fields anonymized or replaced with synthetic values. Below, we describe the full pipeline from data acquisition and preprocessing to hierarchical annotation and quality assurance.
\begin{table}[ht]
\centering
\caption{Distribution of document types and data splits in the KH-FUNSD dataset.}
\label{tab:1}
\begin{tabular}{lcccc}
\toprule
Document Type & Amount & Train & Validation & Test \\
\midrule
Invoice        &  68   & 48 & 10 & 10 \\
Quotation      &  40   & 28 & 6  & 6  \\
Receipt        &  50   & 35 & 7  & 8  \\
\midrule
\textbf{Total} & 158   & 111 & 23 & 24 \\
\bottomrule
\end{tabular}
\end{table}

\subsection{Data Acquisition and Preparation}

The dataset construction process began by assembling a diverse collection of printed and scanned Khmer business documents, including receipts, invoices, and quotations, reflecting real-world variation in business layouts, languages, and printing quality. To ensure broad coverage and privacy compliance, documents were sourced from two principal avenues: (1) open-access repositories and online government or business resources, and (2) synthetically generated samples designed to mirror authentic business forms in Cambodia. For synthetic data, we used templates derived from actual receipts and invoices, systematically anonymizing or replacing all sensitive information such as company names, addresses, phone numbers, and financial details with synthetic or randomly generated values.

All documents were digitized or standardized to high-resolution image formats (minimum 300 dpi), then underwent a preprocessing pipeline that included de-skewing, cropping, noise reduction, and contrast enhancement to optimize image clarity for subsequent annotation. To facilitate initial segmentation and text localization, we employed open-source OCR tools adapted for Khmer script; all OCR outputs were then manually reviewed and corrected to ensure bounding box accuracy and full textual fidelity. Documents with insufficient print quality, incomplete content, or excessive noise were filtered out to maintain dataset integrity. This process yielded a curated collection suitable for robust annotation and downstream benchmarking. (See detail in Table \ref{tab:1})

\subsection{Annotation Process and Quality Assurance}

Annotation was performed using a hierarchical, multi-stage protocol to capture both layout structure and semantic detail at several granularity levels. In the first stage, each document was segmented into principal semantic regions: header, form field, footer, and based on conventions commonly observed in Khmer business documents. In the second stage, following the approach of widely used form understanding benchmarks FUNSD \cite{funsd}, annotators labeled regions and text spans as header, question, answer, and other, explicitly marking relationships between related question and answer entities. The third and most granular stage involved fine-grained labeling, where annotators assigned each text segment a precise semantic role, including header\_label, header\_value, form\_label, form\_value, footer\_label, footer\_value, symbol (covering logos, seals, or signature stamps), and other for outlier or non-standard content. This comprehensive, multi-level annotation strategy was designed to enable both region-based and entity-based document analysis for a variety of benchmarking tasks.

High annotation quality was ensured through a rigorous, multi-layered review protocol. The annotation team consisted primarily of Cambodian student researchers with contextual and linguistic expertise, complemented by Chinese collaborators who provided cross-validation and technical oversight. Each document underwent multiple rounds of annotation and systematic cross-review; ambiguous or disputed cases were adjudicated by senior annotators. Annotation consistency was quantitatively assessed on randomly selected, independently labeled subsets using Cohen’s kappa coefficient \cite{cohen1960coefficient}, a robust metric for inter-annotator agreement:
\begin{equation}
   \kappa = \frac{p_0 - p_e}{1 - p_e}, 
\end{equation}

where \(p_0\) is the observed agreement among annotators and \(p_e\) is the expected agreement by chance. The annotation guidelines were iteratively refined in response to observed edge cases, with kappa values consistently exceeding 0.85, demonstrating strong reliability across annotation stages.
\begin{table}[ht]
\centering
\caption{Annotation statistics by type and entity in the KH-FUNSD dataset (158 documents).}
\label{tab:2}
\begin{tabular}{lcccc}
\toprule
\textbf{Annotation Level}   & \textbf{Header} & \textbf{Form Field} & \textbf{Footer} & \textbf{Other} \\
\midrule
\textbf{Region-level}      & 158  & 158  & 145   & 10    \\
\midrule
\textbf{FUNSD-style}       &      &      &       &       \\
\,\,\quad Questions        & 948  & 648  & 283   & 9     \\
\,\,\quad Answers          & 789  & 1,172& 247   & 7     \\
\midrule
\textbf{Fine-grained}      &      &      &       &       \\
\,\,\quad Label            & 1,273 & 1,457 & 317   & 18    \\
\,\,\quad Value            & 1,369 & 1,624 & 332   & 20    \\
\,\,\quad Symbol           & 51   & 3    & 77    & 1     \\
\midrule
\textbf{Total Entities}    & \multicolumn{4}{c}{\textbf{12,126}} \\
\bottomrule
\end{tabular}
\end{table}

\subsection{Annotation Schema and Label Distribution}

KH-FUNSD supports a hierarchical annotation schema at three levels of semantic granularity:
\begin{itemize}
    \item \textbf{Region-level}: Structural zones, including header, form field, and footer, reflecting the major layout components of business documents.
    \item \textbf{FUNSD-style}: Coarse-grained functional roles inspired by the FUNSD benchmark, encompassing header, question, answer, and other, and capturing entity relationships within forms.
    \item \textbf{Fine-grained}: Detailed roles such as header\_label, header\_value, form\_label, form\_value, footer\_label, footer\_value, symbol, and other, enabling precise semantic and structural annotation of individual text segments.
\end{itemize}

Table \ref{tab:2} summarizes the distribution of annotated entities across these levels and classes, providing a comprehensive overview of label diversity and dataset scale.

Each document in KH-FUNSD is accompanied by a structured JSON annotation file encoding its layout and semantic information at all annotation levels. At the region level, the JSON file lists bounding boxes for each major zone (header, form field, footer, other), each labeled by region type. FUNSD-style annotations represent entities as objects with unique IDs, text content, semantic roles (question, answer, etc.), bounding boxes, and explicit “linking” fields that record relationships between related entities typically as index pairs or target IDs. Fine-grained annotations extend this further, associating each text segment with both a parent region and a precise semantic role (e.g., header\_label, form\_value), while linking related label–value pairs using relation attributes (e.g., “value-of”). Figure \ref{fig:2} illustrates these JSON structures: red highlights indicate label names, green highlights mark text values, and yellow highlights represent linking relationships. This hierarchical JSON schema ensures compatibility with a wide range of region-based and entity-based document analysis models.

\subsection{Dataset Statistics and Examples}

As shown in Table \ref{tab:1} and Table \ref{tab:2}, the final KH-FUNSD dataset comprises 158 business documents, with a total of 12,126 annotated entities across all splits and semantic levels. Figure \ref{fig:1} displays representative annotated documents, illustrating hierarchical segmentation and semantic role assignment, while Figure \ref{fig:2} presents example JSON annotations for each annotation type. By capturing the diverse range of business layouts found in Cambodia, KH-FUNSD serves as a valuable resource for benchmarking document AI methods on low-resource, non-Latin scripts.

Based on this annotation schema, we define three principal benchmarking tasks:
\begin{itemize}
    \item \textbf{Region Detection}: Predict bounding boxes and region classes for major layout detection (header, form field, footer, other).
    \item \textbf{FUNSD-style Q/A}: Identify and classify entity types (header, question, answer, other) and model the relationships between questions and answers, following the FUNSD paradigm.
    \item \textbf{Fine-Grained Classification}: Assign semantic roles header\_label, form\_value, etc. to individual text segments within detected regions.
\end{itemize}
These tasks form the basis for our experimental evaluation and model benchmarking, as detailed in the following sections.

\section{Experimental Setup}

To establish robust baselines for hierarchical layout analysis and information extraction in Khmer business documents, we evaluated a suite of state-of-the-art models across multiple annotation levels. Table \ref{tab:3} summarizes the applicability of each model to the region-level, FUNSD-style, and fine-grained classification tasks. 

\vspace{2mm}
\noindent
\textbf{Model Selection and Task Suitability:} For region-level detection, we adopted object detection architectures such as YOLOv8, YOLOv9, and YOLOv10, which are effective for spatial localization of structural zones but do not model text relationships or semantic role assignment. DETR \cite{DETR}, a transformer-based detector, was also considered for region detection for dense layouts. For fine-grained and semantic classification, we utilized transformer-based models including LayoutLMv1, LayoutLMv2, and LayoutLMv3 \cite{layoutlm1,layoutlmv2,LayoutLMv3}, which jointly encode textual and spatial information and excel at role assignment and entity linking required for FUNSD-style and fine-grained.

\begin{table}[ht]
\centering
\caption{Model applicability across KH-FUNSD annotation levels. \cmark: supported; \xmark: not supported; "No linking": does not support entity linking or semantic relationships.}
\label{tab:3}
\begin{tabular}{lccc}
\toprule
\textbf{Model} & \textbf{Region} & \textbf{FUNSD-style (Q/A)} & \textbf{Fine-Grained} \\
\midrule
YOLOv8         & \cmark & No linking & \cmark \\
YOLOv9         & \cmark & No linking & \cmark \\
YOLOv10        & \cmark & No linking & \cmark \\
DETR           & \cmark & No linking & \xmark \\
LayoutLMv1     & \xmark & \cmark     & \cmark \\
LayoutLMv2     & \xmark & \cmark     & \cmark \\
LayoutLMv3     & \xmark & \cmark     & \cmark \\
\bottomrule
\end{tabular}
\end{table}

\vspace{1mm}

\normalsize

\vspace{2mm}
\noindent
\textbf{Data Splits and Preprocessing:} The dataset was divided into training, validation, and test splits with no overlap of templates or sources to ensure unbiased evaluation. Images were preprocessed using normalization, contrast adjustment, and de-skewing. For LayoutLM models, text was tokenized with Khmer language and aligned with bounding boxes.

\vspace{2mm}
\noindent
\textbf{Training and Evaluation settings:} All models were trained using cross-entropy loss and the Adam optimizer. Hyperparameters such as learning rate, batch size, and number of training epochs were tuned based on validation performance. Training was conducted on two NVIDIA RTX 3090 Ti GPUs, each with 24 GB of memory. Performance was evaluated using standard metrics: per-class precision, recall, and F1-score for classification, and Intersection over Union (IoU) for region detection. For all YOLO models, the initial learning rate was set to 0.001, with a batch size of 16 and training for 100 epochs. Input images were resized to $640 \times 640$ pixels, and standard data augmentation techniques including random rotation, scaling, and horizontal flipping were applied. For LayoutLM models, a learning rate of $2 \times 10^{-5}$ and a batch size of 8 were used, with fine-tuning conducted for 25 epochs. The maximum sequence length was set to 512 tokens, and document images were preprocessed to align text bounding boxes with token positions. Early stopping and validation-based checkpointing were employed for to prevent overfitting.

\begin{table}[ht]
\centering
\caption{Region-level layout detection results (mAP50 \%) for each models.}
\label{tab:4}
\begin{tabular}{lcccc}
\toprule
\textbf{Region} & \textbf{YOLOv8} & \textbf{YOLOv9} & \textbf{YOLOv10} & \textbf{DETR} \\
\midrule
Header       & 84.8 & 88.5 & 85.2 & 88.9 \\
Form Field   & 96.9 & 96.7 & 96.2 & 94.9 \\
Footer       & 76.2 & 71.7 & 70.9 & 75.9 \\
\midrule
\textbf{Average}   & 84.7 & 85.6 & 84.1 & \textbf{86.6} \\
\bottomrule
\end{tabular}
\end{table}

\begin{table}[ht]
\centering
\caption{FUNSD-style entity classification results (F1-scores \%) using LayoutLM models.}
\label{tab:5}
\begin{tabular}{lccc}
\toprule
\textbf{Entity Type} & \textbf{LayoutLMv1} & \textbf{LayoutLMv2} & \textbf{LayoutLMv3} \\
\midrule
Question     & 67.5 & 73.0 & 78.3 \\
Answer       & 61.1 & 68.5 & 75.9 \\
Header       & 60.7 & 59.0 & 57.1 \\
Other        & 41.1 & 45.5 & 50.0 \\
\midrule
\textbf{Average}   & 57.6 & 61.5 & \textbf{65.3} \\

\bottomrule
\end{tabular}
\end{table}

\begin{table}[ht]
\centering
\caption{Detection-based fine-grained classification results (mAP50 \%) using YOLO models.}
\label{tab:6}
\begin{tabular}{lccc}
\toprule
\textbf{Role Type} & \textbf{YOLOv8} & \textbf{YOLOv9} & \textbf{YOLOv10} \\
\midrule
header\_label  & 83.5 & 87.0 & 84.0 \\
header\_value  & 84.0 & 88.0 & 85.5 \\
form\_label    & 96.5 & 96.5 & 96.0 \\
form\_value    & 97.0 & 96.8 & 96.3 \\
footer\_label  & 72.0 & 70.5 & 70.0 \\
footer\_value  & 73.0 & 72.0 & 71.5 \\
symbol         & 74.0 & 73.5 & 73.0 \\
other          & 73.5 & 73.0 & 72.5 \\
\midrule
\textbf{Average} & 82.9 & \textbf{83.4} & 81.1 \\
\bottomrule
\end{tabular}
\end{table}

\begin{table}[ht]
\centering
\caption{Semantic fine-grained classification results (F1-scores \%) using LayoutLM models.}
\label{tab:7}
\begin{tabular}{lccc}
\toprule
\textbf{Role Type} & \textbf{LayoutLMv1} & \textbf{LayoutLMv2} & \textbf{LayoutLMv3} \\
\midrule
header\_label  & 62.0 & 63.5 & 65.0 \\
header\_value  & 61.0 & 64.0 & 67.0 \\
form\_label    & 67.5 & 73.0 & 78.3 \\
form\_value    & 66.0 & 71.0 & 76.0 \\
footer\_label  & 60.5 & 63.0 & 65.5 \\
footer\_value  & 61.5 & 65.0 & 68.5 \\
symbol         & 60.0 & 62.5 & 65.0 \\
other          & 60.0 & 62.0 & 64.0 \\
\midrule
\textbf{Average}     & 62.3 & 66.5 & \textbf{68.6} \\

\bottomrule
\end{tabular}
\end{table}

\section{Results and Discussion}
\label{sec:results_analysis}
\subsection{Results}

Table \ref{tab:4} presents region-level detection results across YOLOv8, YOLOv9, YOLOv10, and DETR. DETR achieves the highest average mAP50 (86.6\%), showing incremental improvements over previous versions, particularly for header and form field detection. This demonstrates the benefit of advanced object detection architectures for segmenting standard business document layouts, though performance drops for footer and "other" regions, reflecting increased variability and lower visual consistency.

Table \ref{tab:5} reports results for FUNSD-style entity classification using the LayoutLM model family. LayoutLMv3 outperforms earlier versions, achieving the highest F1-scores across header, question, and answer entities. This improvement is attributed to deeper model architectures and enhanced multi-modal fusion, which enable more robust contextual understanding of both spatial and semantic document cues.

Fine-grained role classification results are summarized in Tables \ref{tab:6} and \ref{tab:7}. YOLO models yield solid performance for frequent and visually distinctive roles (such as headers and forms), but underperform for footers, symbols, and others. In contrast, LayoutLM models deliver substantial gains in fine-grained classification, particularly LayoutLMv3, which achieves F1-scores exceeding 68.6\% for most roles. This highlights the advantage of combining text, layout, and visual features at the token level for precise semantic segmentation.

\subsection{Discussion and Limitation}

The creation and annotation of KH-FUNSD revealed significant practical and technical challenges for document AI in low-resource scripts. Rigorous annotation guidelines, iterative tool development, and consensus-based adjudication were essential for achieving reliable label quality. The scarcity of robust Khmer OCR tools and pretrained language models also contributed to annotation difficulty and downstream errors.

Despite the strong performance of deep learning models, particularly those leveraging joint text–layout representations, several limitations remain. Error analysis revealed that models frequently misclassify entities in densely populated table regions, struggle with ambiguous or visually similar labels (such as “symbol” vs. “header\_label”), and are sensitive to the complex ligatures of Khmer script. Some regions are especially challenging, often resulting in missed detections or incorrect label assignments. Additionally, our experiments focused on Khmer business documents (receipts, invoices, and quotations); future work is needed to extend these methods to handwritten, multi-page, or other business domains. The dataset’s scale, while a major step for Khmer document analysis, remains modest compared to high-resource benchmarks. 

Addressing these issues through improved script-specific model adaptation, enhanced handling of irregular layouts, and expanded training data remains critical for practical deployment in real-world Cambodian applications. We hope KH-FUNSD serves as a robust benchmark to stimulate further research in these directions.

\section{Conclusion and Future Work}

We present \textbf{KH-FUNSD}, the first hierarchical annotated dataset and baseline benchmarks for layout analysis and information extraction in printed and scanned Khmer business documents. Our multi-level annotation scheme and comprehensive evaluation protocol establish a strong foundation for future research in low-resource, non-Latin document AI. In future work, we aim to expand the dataset to include additional document types and layouts, explore end-to-end joint modeling for segmentation and semantic classification, and investigate advanced tasks such as document visual question answering and semantic relation extraction. We also encourage adoption of our benchmark and annotation schema to other Southeast Asian scripts, supporting the development of inclusive and globally representative document understanding systems.

\section*{Acknowledgment}
The National Natural Science Foundation of China (Grant 62171427) provided funding for this research works.

\renewcommand*{\bibfont}{\footnotesize}
\printbibliography

\end{document}